\documentclass[letterpaper, 10 pt, conference]{ieeeconf}  

\IEEEoverridecommandlockouts                              

\usepackage{cite}
\usepackage{amsmath,amssymb,amsfonts}
\usepackage{graphicx}
\usepackage{textcomp}
\usepackage{xcolor}
\usepackage{bm}
\usepackage{graphicx}
\usepackage{pifont}
\usepackage{amssymb}
\usepackage[hidelinks]{hyperref}
\usepackage{siunitx}

\usepackage{booktabs}

\usepackage{algorithm}
\usepackage{algpseudocode}

\usepackage{graphicx} 
\usepackage{epstopdf}

\usepackage{soul}

\definecolor{YelloGreen}{RGB}{173,255,47}

\usepackage{float}

\usepackage{caption}

\title{\LARGE \bf
CubeDN: Real-time Drone Detection in 3D Space from Dual mmWave Radar Cubes}

\author{Yuan Fang$^{1}$, Fangzhan Shi$^{1}$, Xijia Wei$^{2}$, Qingchao Chen$^{3*}$, Kevin Chetty$^{1*}$ and Simon Julier$^{2}$
\thanks{$^{1}$Yuan Fang, Fangzhan Shi, Kevin Chetty is with Department of Security and Crime Science, University College London, United Kingdom}
\thanks{$^{2}$Xijia Wei, Simon Julier is with Department of Computer Science, University College London, United Kingdom}
\thanks{$^{3}$Qingchao Chen is with the National Institute of Health Data Science, Institute of Medical Technology, and the State Key Laboratory of General Artificial Intelligence, Peking University, China. {QC's work was supported by the National Natural Science Foundation of China
under Grant (62201014), Peking University Medicine Seed Fund for Interdisciplinary Research (PKU2024LCXQ028), and the Fundamental Research Funds for the Central Universities.}}
\thanks{$^{*}$ Co-corresponding author}
}

\begin{document}

\maketitle
\thispagestyle{empty}
\pagestyle{empty}

\begin{abstract}


As drone use has become more widespread, there is a critical need to ensure safety and security. A key element of this is robust and accurate drone detection and localization. While cameras and other optical sensors like LiDAR are commonly used for object detection, their performance degrades under adverse lighting and environmental conditions. Therefore, this has generated interest in finding more reliable alternatives, such as millimeter-wave (mmWave) radar. Recent research on mmWave radar object detection has predominantly focused on 2D detection of road users. Although these systems demonstrate excellent performance for 2D problems, they lack the sensing capability to measure elevation, which is essential for 3D drone detection. To address this gap, we propose CubeDN, a single-stage end-to-end radar object detection network specifically designed for flying drones. CubeDN overcomes challenges such as poor elevation resolution by utilizing a dual radar configuration and a novel deep learning pipeline. It simultaneously detects, localizes, and classifies drones of two sizes, achieving decimeter-level tracking accuracy at closer ranges with overall $95\%$ average precision (AP) and $85\%$ average recall (AR). Furthermore, CubeDN completes data processing and inference at 10Hz, making it highly suitable for practical applications.

\end{abstract}

\section{INTRODUCTION}

In an era of smart and autonomous robotics, drones are increasingly used in various consumer, industrial, and research applications. Tracking and detecting flying drones has become a key issue, as precise 3D positioning is essential for autonomous flying, localization, and navigation. Additionally, the rise in illegal drone activity poses potential threats to airspace distribution, privacy, and security. As a result, the demand for effective drone detection and classification is rising~\cite{drone_intro}.

Non-cooperative detection is ideal in situations when no modifications are made to the drone. Cameras are often the primary sensors used, enabling deep learning-based object detection and classification techniques in the field of computer vision, making drone detection an increasingly focused area of research~\cite{cvdrone2}. However, cameras and other optical sensors including LiDAR, depth cameras and thermal cameras are likely to suffer from visual degradation in adverse environments including poor illumination, bad weather, smoke and dust, since they require a clear line-of-sight to function effectively\cite{millimap}. 
Furthermore, most systems use 2D images and can only estimate object depth from bounding box size. Such depth estimates are extremely inaccurate, making trajectory estimation in 3D space extremely challenging~\cite{monocamera}.



mmWave radar has recently gained widespread adoption for object detection due to its wireless sensing capabilities and robustness in adverse environments. While radar has traditionally been used in aviation and military applications for tracking aircraft over long distances, there is a growing trend to leverage mmWave sensing technology for drone detection \cite{echodyne}, but many existing solutions are still large, stationary, high-power, customized, costly, and primarily designed for defense or critical infrastructure security.

Meanwhile, the latest semiconductor fabrication have enabled the development of low-cost (\textless\$350) and lightweight single-chip mmWave radars, with some even integrating the antenna into the chip \cite{1843aop}. This is a significant improvement compared to the bulky mechanically scanning mmWave radar \cite{navtech}, leading to broader more applications in the fields of smart robotics and autonomous system, including localization and mapping \cite{millimap,radarhd,radarize}, sensing \cite{mBeat} and tracking \cite{mid}. Such low-cost and lightweight mmWave radars can easily be integrated into mobile robot platforms \cite{batmobility} and handheld devices \cite{milliego}, greatly broadening the application of mmWave sensing across various fields.


Consequently, mmWave radar object detection has become a popular research topic, with many effective frameworks \cite{vehicledetection,cnnradarcube,rodnet, raddet, rampcnn} proposed for detecting ground-based objects like pedestrians, cyclists, and vehicles, where height is often ignored due to the poor elevation resolution of single-chip radar. Since elevation and height information is not required, they perform 2D detection similar to Bird’s Eye View (BEV). In contrast, drones operate in 3D space, requiring accurate altitude estimation. Moreover, drones are also much smaller than typical road users (e.g., our small drone is only \SI{250}{\milli\metre} diagonally), making detection even more challenging. Additionally, classifying drones of different sizes (small and large) adds complexity, as their size difference is less pronounced than that of road users.

Overall, drone detection with single-chip mmWave radar faces challenges like poor elevation estimation, long-range detection difficulties, classification uncertainty, and the need for real-time processing and accurate tracking. To overcome these challenges, we propose CubeDN (Cube Drone DetectioN), the first mmWave radar detection system specifically designed for drones, to the best of our knowledge. CubeDN can detect the presence of drones, classify their types (small or large), and predict their positions in 3D space. The CubeDN system employs a non-cascaded dual radar configuration to enhance elevation resolution and provide robust performance in both detection and localization. It achieves this by leveraging a deep learning pipeline to automatically learn complex features that are intractable for handcrafted parameters while maintaining real-time performance.

Contributions:

\begin{itemize}

\item We propose a novel mmWave radar drone detection network called CubeDN, which achieves state-of-the-art performance in both detection and localization by accurately estimating drones' position and classifying their types.

\item We introduce an innovative method for enhancing elevation estimation in 3D space using two non-cascaded radars, along with a novel cube fusion technique that preserves critical correlations between range, doppler, azimuth, and elevation for object detection.

\item We demonstrate the truly real-time detection capabilities that CubeDN can process and infer within 100ms (10Hz) after acquiring radar raw data, enabled by CUDA acceleration, making it highly suitable for real-world applications.


\end{itemize}


\section{Related Work}

Detecting drones in the sky presents unique challenges due to their small size and other difficult-to-handle characteristics. To address these challenges, researchers have developed advanced computer vision algorithms \cite{cvdrone}, making drone detection a specialized research field with dedicated datasets \cite{cvdrone_dataset}. However, depth estimation remains an issue, as most detection methods rely on 2D images. Beyond cameras, sensors like LiDAR, which actively measures range, have also been used for drone detection \cite{lidardrone}, but this approach typically requires multi-line LiDAR and is still susceptible to adverse weather. Acoustic detection methods using microphone arrays are also vulnerable to background noise and have limited range\cite{acoustic_detection, acoustic_detection2}. Additionally, capturing wireless signals transmitted by drones during teleoperation can only detect their presence without effective tracking \cite{RFdrone}, and is likely to experience interference from other signals.

Radar-based drone detection has also been extensively researched, with most studies focusing on pulse radar \cite{pulseradar}, continuous wave radar \cite{cwradar}, and X-Band radar \cite{xband}. Meanwhile, mmWave radar is emerging as a promising alternative due to its higher range and doppler resolution, offering advantages for object detection. However, its application in drone detection is still relatively unexplored. Existing studies using mmWave radar often face similar limitations to pulse and continuous wave radars, such as only detecting drone presence \cite{detect_only} without classification or 3D position estimation \cite{35Gdrone}, requiring special testing environments for classification\cite{droneclassification}, or unable to track multiple drones simultaneously\cite{mdrone}.

Meanwhile, mmWave radar object detection has become a popular research topic, especially with the application of deep learning to radar signal processing \cite{vehicledetection,cnnradarcube,rodnet, raddet, rampcnn}. These techniques can overcome challenges in conventional methods (e.g. reliance on handcrafted features), providing enhanced accuracy and robustness in diverse scenarios. However, most research still focuses on road users detection for driver assistance, using radar point clouds \cite{pointcloud_detect} or raw data as input while less projects have incorporated elevation into the detection process, and only some have achieved this with cascaded radar \cite{cubecrop}. The most similar work to that presented in this study is mDrone\cite{mdrone}, which also uses deep learning and raw data to localize drones in 3D space. However, it lacks the ability to detect the presence of drones, classify them, or support multiple drones simultaneously. In summary, Table I highlights CubeDN's ability to detect two drone types in 3D space, compared with other mmWave detection methods.

\begin{table}[]
\scriptsize 
\setlength{\tabcolsep}{4pt} 
\begin{tabular}{@{}lcccc@{}}
\toprule
Methods                                                    & Input                                                               & Objects                                                          & Multi-Obj & 3D  \\ \midrule
RODNet\cite{rodnet}                                                     & RA                                                                  & Road Users                                                       & Yes       & No  \\
RAMP-CNN\cite{rampcnn}                                                   & RA, RD, DA                                                          & Road Users                                                       & Yes       & No  \\
RADDet\cite{raddet}                                                   & RAD Cube                                                            & Road Users                                                       & Yes       & No  \\
\begin{tabular}[c]{@{}l@{}}Carry \\ Detection \cite{cubecrop}\end{tabular} & \begin{tabular}[c]{@{}c@{}}RAE Cube\\ (Cascaded Radar)\end{tabular} & \begin{tabular}[c]{@{}c@{}} Human with (concealed) \\ Knife/Phone/Laptop\end{tabular}   & No       & Yes \\
mDrone \cite{mdrone}                                                     & RA, RE                                                              & Drone                                                            & No        & Yes \\ \midrule
\begin{tabular}[c]{@{}l@{}}CubeDN\\ (Ours)\end{tabular}    & \begin{tabular}[c]{@{}c@{}}RAD + RED \\ Cube\end{tabular}           & \begin{tabular}[c]{@{}c@{}}Drones\\ (Small + Large)\end{tabular} & Yes       & Yes \\ \bottomrule
\end{tabular}

\captionsetup{justification=centering, singlelinecheck=false}
\captionsetup{width=\textwidth}
\caption{Comparison of mmWave object detection methods.}
\vspace{-0.6cm}
\end{table}






\section{Radar Data Processing and Dual Radar fusion}

\subsection{Radar Data Format and Cube Representation}
mmWave radar can measure objects' range, angle and doppler velocity. The typical method for processing returned mmWave radar signals is to use Fast Fourier Transform (FFT) to extract into a Range-Angle-Doppler (RAD) data cube, a 3D tensor representation that contains the information about objects’ range, angle and velocity as shown in Fig. 2a. While some projects \cite{raddet,cnnradarcube} take the full RAD cube as input for detection, many others \cite{ra_classification,rd,batmobility} compress the cube into heatmaps by selecting any two dimensions in the 3D RAD cube and summing along the third, resulting in Range-Angle (RA), Range-Doppler (RD), and Doppler-Angle (DA) heatmaps as shown on the corresponding sides of the RAD cube in Fig. 2.a. However, compression from 3D RAD Cube into 2D heatmaps breaks the correlation between the range, doppler and angle, leading to information loss, especially some projects only take RA for detection \cite{rodnet} and discard RD or DA. Although 2D heatmaps are easier for image-based networks to process compared to the 3D cube, using all three heatmaps (RA, RD, DA) requires separate feature extractors which adds more complexity. \cite{rampcnn,vehicledetection}. To avoid this and retain full data integrity, we keep the original 3D RAD cube format, preserving critical and correlated information like reflection amplitude, doppler patterns, and angle of arrival of the objects.


\subsection{Dual Radar Fusion and Preprocessing}



The radar we use features a 3 Tx and 4 Rx MIMO antenna array, forming 12 virtual antennas (VA) from Tx-Rx pair combinations, with 8 VA for azimuth estimation but only 2 VA for elevation \cite{mimo_radar}, as illustrated in the antenna layout shown in Fig. 2a. This results in a much lower elevation resolution (\SI{60}{\degree}) compared to the azimuth resolution (\SI{15}{\degree}). To accurately detect drones or other objects in 3D space, the elevation resolution needs to be higher, ideally matching the azimuth resolution for consistency. To achieve this, we use two identical radars, enabling only the azimuth antennas on both, and rotating one of them by 90 degrees. This rotation allows the radar that originally provided azimuth information to instead provide elevation information.



Considering both radars are on the same plane and the 5 cm gap between them is smaller than the range resolution of 11.6 cm, the range and Doppler dimensions from both radars are expected to be theoretically identical. Slight practical differences may arise due to factors like calibration errors,  environmental influences and noise. These differences are expected to be minimal and common-mode, consistent across both radars. The azimuth and elevation measurements, however, are different.

The individual outputs of the horizontal and vertical radars after FFT are a Doppler-Range-Azimuth cube and a Doppler-Range-Elevation cube, represented as $\bm{H} \in \mathbb{R}^{D \times R \times A}$ and $\bm{V} \in \mathbb{R}^{D \times R \times E}$, respectively. To facilitate their fusion, an additional  dimension is introduced to each data cube, resulting in the expanded forms $\bm{H^{\prime}} \in \mathbb{R}^{D \times R \times A \times 1}$ and $\bm{V^{\prime}} \in \mathbb{R}^{D \times R \times 1 \times E}$. Given that the doppler and range dimensions are identical between $\bm{H^{\prime}} $ and $\bm{V^{\prime}}$, an element-wise multiplication can be performed across the azimuth and elevation dimensions, resulting in a four-dimensional data cube, $\bm{Cube\_4D}$ that each element value is calculated as:

\vspace{-0.1cm}

\begin{equation}
\mathbf{Cube\_4D}[h, i, j, k] = \bm{H^{\prime}}[h, i, j, 1] \times \bm{V^{\prime}}[h, i, 1, k]
\end{equation}

$\bm{Cube\_4D} \in \mathbb{R}^{D \times R \times A \times E}$ fuses the azimuth and elevation information from the horizontal and vertical radars while preserving the doppler and range information in a single 4D Cube. The multiplicative fusion effectively captures the relationship between azimuth and elevation across two radars, enhancing strong signal co-occurrence where objects exist in 3D space while suppressing background noise, all without the added complexity of using network to separately extract and concatenate features from each radar.


\section{CubeDN: Drone Detection From Radar Cube }

\addtocounter{figure}{+1}
\begin{figure*}[thbp]
\centering
\includegraphics[width=1.0\textwidth]{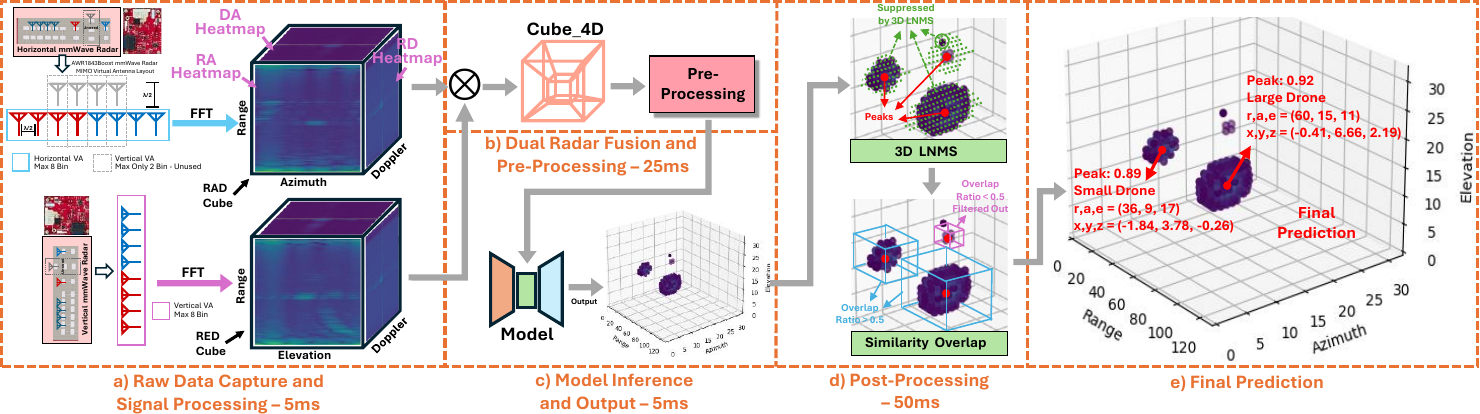}
\vspace{-0.5cm}
\caption{The processing workflow for a single frame, from acquiring the radar data to producing the final prediction (a-e).}
\label{1}
\vspace{-0.5cm}
\end{figure*}

\subsection{CubeDN Network architecture}
\vspace{-0.2cm}
\addtocounter{figure}{-2}
\begin{figure}[H]
\centering
\includegraphics[width=0.45\textwidth]{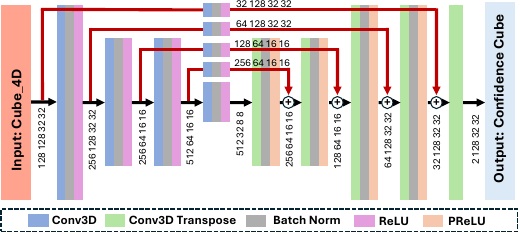}
\vspace{-0.1cm}
\caption{The network architecture of CubeDN with input and output dimensions of each layer.}
\label{1}
\vspace{-0.3cm}
\end{figure}

CubeDN employs a 3D CNN encoder-decoder architecture, as shown in Fig. 1. Skip connections are added between corresponding encoding and decoding feature maps. To ensure that the feature dimensions match for element-wise addition, an additional 3D convolution is applied within the skip connections. The final layer is a classification head that reduces the channel size to the number of object classes.

After passing through the network, the input $\bm{Cube\_4D}$ with dimension $(D,R,A,E)$ will have the output prediction $\bm{\hat{Q}}$, a confidence cube with dimension $(C_{cls}, R, A, E)$, where $C_{cls}$ represents the number of object classes. In this project, $C_{cls}=2$ for detecting small and big drones. Notably, the final prediction does not provide doppler estimation, indicating that the network implicitly utilizes doppler information to aid in classification.


CubeDN predicts separate confidence cubes $\bm{\hat{Q}}_{cls}$ for each object class, and within each confidence cube, it can identify multiple objects of the same class. These predicted confidence cubes $\bm{\hat{Q}}_{cls}$ represent the likelihood of objects appearing at specific locations in the range-azimuth-elevation space, just as we define in the ground truth $\bm{Q_{cls}}$. By minimizing the difference between the predicted and ground truth confidence cubes, the network learns to accurately predict both the class and location of objects. Since the confidence values are continuous probabilities rather than discrete labels, the network is trained using the Mean Squared Error (MSE) loss function:

\begin{equation}
\ell = \frac{1}{{Cls}} \sum_{{cls=1}}^{Cls} \frac{1}{N} \| \bm{Q}_{cls} - \bm{\hat{Q}}_{{cls}} \|_2^2
\end{equation}

\noindent 
where $\bm{Q}_{cls}$ represents the ground truth and $\bm{\hat{Q}}_{cls}$ is the prediction confidence cube for each class. $Cls$ is the total number of object classes; $N = R \times A \times E$ is the total number of elements in the confidence cube $\bm{Q}_{cls}$ or $\bm{\hat{Q}}_{cls}$.

\subsection{Ground Truth Generation for Training}

The na\"ive ground truth is the drone position $(x,y,z)$ in Cartesian coordinate that needs to be converted into  $\bm{p}=(r,a,e)$ under radar Polar coordinate, but $\bm{p}$ cannot be directly used for training, because CubeDN predicts a confidence cube rather than a coordinate point. The input to the network, after excluding the Doppler dimension, is a Range-Azimuth-Elevation (RAE) cube that reflects the amplitude of the object’s reflection. Since the size and shape of the object influence the extent of the reflection in the azimuth and elevation dimensions, a larger drone will occupy a broader area in these dimensions, while a smaller drone will occupy a smaller area. Similarly, to construct an appropriate ground truth confidence cube for each class $\bm{Q_{cls}}$, we assign the maximum confidence value to the element at the location $\bm{p}=(r,a,e)$, where the drone is located, while the surrounding area should have progressively lower confidence values than the maximum point $\bm{p}$. This reflection decay distribution is modeled using a Gaussian function, where the confidence value of each element $conf_{i,j,k}$ in the ground truth confidence cube $\bm{Q_{cls}}$ is calculated as:

\begin{equation}
\begin{gathered}
conf_{i,j,k} = k \cdot \exp\left(-\frac{d^2}{2\sigma^2}\right) \\
d = \sqrt{(i-r)^2 + (j-a)^2 + (j-e)^2}
\end{gathered}
\end{equation}

\noindent
where $d$ is the Gaussian decay distance between each element $(i,j,k)$ in the cube $\bm{Q_{cls}}$ to the drone position $\bm{p}=(r,a,e)$. Empirically, we set $\sigma = 1$ for small drone and $\sigma = 2$ for large drone. The factor $k$ is set to 1, ensuring the element at the drone position $\bm{p}$ will have the maximum confidence value 1.0. Eventually, We filter out elements with confidence values lower than 0.05 by setting them to 0, resulting the final $\bm{Q_{cls}}$. The set of elements with confidence values above 0.05 is referred to as the ground truth Gaussian mask $\bm{G}$.

\subsection{Post-processing for Model Predictions}

After prediction, the confidence cube $\bm{\hat{Q}}_{cls}$ requires post-processing to find potential object locations and filter out any outliers, including 3D LNMS and similarity comparison as shown in the Fig. 2.d.

\paragraph{3D LNMS (Location Non-Maximum Suppress)}

To identify potential object locations in the confidence cube $\bm{\hat{Q}}_{cls}$, we apply a 3D Location-based Non-Maximum Suppression (NMS) to filter out redundant candidates while only maintaining the peak elements. This process involves sorting the elements by confidence, selecting the highest peak, and then suppressing nearby elements within a defined radius, repeating until only distinct peaks remain.

The steps are as follows:

\begin{enumerate}
\item Sort elements in the prediction cube $\bm{\hat{Q}}_{cls}$ by confidence values (greater than 0.01) to form the set  $\mathcal{L}$.
\item Iteratively select the highest confidence element in $\mathcal{L}$ as a potential object location, and apply a larger Gaussian mask ($\sigma = 3$ for small drones and $\sigma = 5$ for large drones) to suppress and remove nearby elements with high confidence from $\mathcal{L}$.
\item Repeat the process until all distinct peaks are identified, resulting in the final detected peaks set $\mathcal{L^{\prime}}$ and their corresponding potential object locations set $ \mathcal{P}$.
\end{enumerate}

\medskip

\paragraph{Similarity and Overlap Ratio} 
The 3D LNMS identifies potential object candidates $\mathcal{L^{\prime}}$ and their locations $\mathcal{P}$ from $\bm{\hat{Q}}{cls}$, but it only performs peak searching without verifying if each peak cluster shape corresponds to a Gaussian mask as in the ground truth. This can result in high-confidence outliers being included, as shown in Fig. 2d. To remove these outliers, we crop a small region around each peak element, producing a $(7,7,7)$ range-azimuth-elevation cube $\bm{\hat{O}}$. By assuming there is an object at this point, we can generate a Gaussian mask $\bm{G}{n}$ using equation (3) and crop a corresponding region $\bm{O}$ from $\bm{G}_{n}$. By comparing the similarity between $\bm{O}$ and $\bm{\hat{O}}$, measured by their overlap ratio, we can determine if the prediction is an outlier. The overlap ratio is empirically set to 0.5 and is calculated as follows:

\vspace{-0.1cm}






\begin{algorithm}
\caption{calculate\_overlap\_ratio}
\begin{algorithmic}
\State \textbf{Input:} $\bm{O}$, $\bm{\hat{O}}$, $\tau$ (threshold, default: 0.05)
\State \textbf{Binarize Arrays:}
\State binary\_$\bm{O}$ $\gets$ 1 if element in $\bm{O}$ $> \tau$, else 0
\State binary\_$\bm{\hat{O}}$ $\gets$ 1 if element in $\bm{\hat{O}}$ $> \tau$, else 0

\State \textbf{Calculate Overlap:}
\State overlap $\gets$ sum of elements where both binary\_$\bm{O}$ and binary\_$\bm{\hat{O}}$ are 1

\State \textbf{Calculate Overlap Ratio:}
\State total\_points $\gets$ sum(binary\_$\bm{O}$) + sum(binary\_$\bm{\hat{O}}$) - overlap

\State overlap\_ratio $\gets$ \textbf{if} total\_points $>$ 0 \textbf{then} overlap / total\_points \textbf{else} 0

\State \textbf{return} overlap\_ratio
\end{algorithmic}
\end{algorithm}

\vspace{-0.35cm}

\section{Implementations}

\subsection{Hardware Design}
\begin{figure}[thbp]
\addtocounter{figure}{1}
\centering
\includegraphics[width=0.48\textwidth]{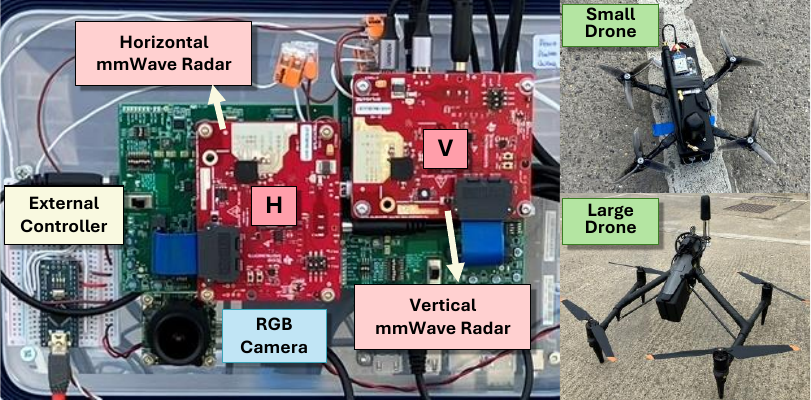}
\caption{Left: Setup of our dual mmWave radar system, horizontally and vertically placed, triggered and synchronized by an external controller. Right: Top: Small 5-inch drone. Bottom: Large DJI Inspire 3 drone.}
\label{1}
\vspace{-0.2cm}
\end{figure}

In this project, we use two identical AWR1843Boost mmWave radars \cite{1843boost} and DCA1000EVM data capture cards \cite{dca1000}, placed horizontally and vertically as shown in Fig. 3 Left. These radars operate independently (not cascaded) but are triggered by the external controller to avoid interference, which also synchronizes them with the camera and other sensors, allowing accurate timestamping of each radar frame.

Two different types of drones are used in our experiments. The small drone is a homemade model, built with a 5-inch frame having a \SI{250}{\milli\metre} diagonal distance (Fig. 3 Right UP), while the large drone is the DJI Inspire 3, a professional heavy-lift cinematic drone with a diagonal distance of \SI{685}{\milli\metre}, as shown in Fig. 3 Right Down. To acquire the drone's position ground truth in 3D space, we use the RTK module for outdoor measurements with an accuracy of \SI{1}{\centi\metre} at \SI{10}{\hertz}, and the Mo-Sys StarTracker for indoor measurements with sub-millimeter accuracy at \SI{60}{\hertz} \cite{stracker}. These sensors are all synchronized to each radar frame.


\subsection{Data Collection and Experiment Setup}

%



\begin{table}[]
\begin{tabular}{@{}lcccc@{}}
\toprule
Drone Type   & Seqs & Raw Frames & Train+Val Frames & Test Frames \\ \midrule
Mini         & 12   & 36K        & $\sim$27K        & $\sim$4K    \\
Large        & 16   & 48K        & $\sim$25K        & $\sim$6K    \\
Mini + Large & 13   & 39K        & $\sim$12K        & $\sim$3K    \\ \midrule
Overall      & 41   & 123K       & $\sim$64K        & $\sim$13K   \\ \bottomrule
\end{tabular}
\captionsetup{justification=centering, singlelinecheck=false}
\caption{Dataset details for CubeDN by drone type.}
\vspace{-0.6cm}
\end{table}

The radar is configured with 255 doppler bins and 256 range bins, but both are compressed to 128 bins to accelerate network computing, while 8 angle bins are padded to 32. This compression reduces the native doppler resolution by half to \SI{0.094}{\metre\per\second} and the range resolution to \SI{0.116}{\metre}, while the maximum detection range (\SI{15}{\metre}) remains unchanged. 

We collect the dataset (Table II) at \SI{10}{\hertz} under various conditions (day/night/windy) across three outdoor car parks, with the drones teleoperated to fly and maneuver randomly in diverse environments. Additionally, we collect 2K frames at an outdoor gate entrance and 600 frames in an indoor office for generalization testing. Frames where the drone hasn't taken off are removed, but 'empty' frames are kept when the drone is beyond the maximum detection range ($>$\SI{15}{\metre}). Frames are also discarded if the ground truth is inaccurate (e.g. RTK accuracy exceeds \SI{2.5}{\centi\metre}). Since CubeDN is trained randomly and doesn't consider time series, ground truth failures are not compensated.




\subsection{Network Training and Real-time Inference}
The model is trained on an RTX 4090 GPU with an i5-12600K and 64GB RAM, taking about 3 days for 70 epochs. Real-time drone detection runs on an RTX 3080Ti laptop with a Ryzen 6900HX and 64GB RAM. Data from two radars transmitted via Ethernet cable is parsed with UDP scripts and decoded using CUDA acceleration, which dramatically reduces the time-consuming operations like FFT from seconds on CPU into just a few milliseconds, enabling real-time performance. The entire process for one frame takes about 90ms, faster than the radar’s 10 Hz frame rate, ensuring timely detection. The detailed time consumption of each module is listed in Fig. 2.


\section{Evaluation Results and Analysis}

We evaluate our drone detection system on the test dataset based on two criteria: drone detection performance in radar coordinates and localization error in Cartesian coordinates. Unlike 2D object detection using bounding boxes, our system predicts the objects' center position in 3D space, where localization accuracy is more critical than detection performance for subsequent tasks like drone tracking and navigation.

\subsection{Drone Detection Results}

\begin{table*}[]
\centering
\begin{tabular}{@{}l|c|ccc|c|ccc@{}}
\toprule
\multicolumn{1}{c|}{Methods}     & AP - Overall & AP$^1$  & AP$^3$  & AP$^5$  & AR - Overall & AR$^1$  & AR$^3$  & AR$^5$  \\ \midrule
CubeCrop \cite{cubecrop} - small drone           & 49.05        & 10.32 & 40.57 & 61.61 & 56.92        & 23.98 & 53.69 & 63.62 \\
CubeCrop \cite{cubecrop} - large drone           & 65.56        & 24.65 & 57.66 & 77.76 & 81.86        & 51.81 & 81.09 & 85.79 \\
CubeCrop \cite{cubecrop} - small + large   drone & 62.17        & 26.72 & 55.87 & 72.19 & 73.99        & 52.64 & 72.52 & 77.71 \\ \midrule
CubeDN \textbf{(Ours)} - small drone             & \textbf{95.25} & 48.62 & 96.55 & 98.87 & \textbf{82.77} & 41.83 & 83.90 & 85.95 \\
CubeDN \textbf{(Ours)} - large drone             & \textbf{96.18} & 48.28 & 97.59 & 99.81 & \textbf{89.13} & 44.74 & 90.45 & 92.49 \\
CubeDN \textbf{(Ours)} - small + large drone     & \textbf{93.32} & 48.84 & 93.34 & 97.99 & \textbf{84.76} & 44.46 & 84.81 & 88.95 \\ \bottomrule
\end{tabular}
\captionsetup{justification=centering, singlelinecheck=false}
\caption{Drone Detection Performance Under Different OLE Thresholds.}
\vspace{-0.6cm}
\end{table*}

{\noindent\textbf{Evaluation Metrics:}} 
As there will be a difference between the predicted drone location and the groundtruth position, we define the evaluation metrics, object localization error (OLE) to represent different localization error tolerance for the detection performance. OLE is calculated by measuring the location errors in the radar coordinate, and the threshold is set to 1, 3, 5 respectively. These thresholds represent the maximum allowable error bins from the predicted position to the ground truth across the Range-Azimuth-Elevation dimensions. OLE can be considered similar to the Intersection over Union (IoU) metric commonly used in 2D image object detection which measures the overlap between the predicted and actual bounding boxes.

{\noindent\textbf{Baseline Methods:}} 
We compare our detection results with the method proposed in \cite{cubecrop} for carried object detection, namely CubeCrop, which follows a 2-stage detection approach that first uses conventional methods including CFAR \cite{cfar} and clustering to detect and identify object candidates from the entire range-azimuth-elevation cube, then crops a small region around each candidate and feeds this cropped area into a classification network. To the best of our knowledge, CubeCrop is the only recent project that explores 3D object detection with mmWave radar using deep learning, though it is originally designed for detecting human carrying with concealed or non-concealed objects. We adapt its 3D detection framework to detect drones using our dataset.

{\noindent\textbf{Results and Findings:}} 
Table III presents the average precision (AP) and average recall (AR) under different OLE thresholds. We evaluated our system in three scenarios: mini drone only, large drone only, and both drones flying together. The results show that CubeDN performs well, achieving $93.32\%$ AP and $84.76\%$ AR when both drones are present. This strong performance is consistent across scenarios with either only the mini or large drone is present, indicating effective detection and classification.

Although performance drops under OLE = 1, this is expected due to the strict localization error tolerance, allowing only a 1-bin difference in range, azimuth, or elevation, leading to many correctly classified results being counted as false negatives. However, performance improves notably to around $94\%$ AP and $85\%$ AR under OLE = 3, which we believe is more representative of real-world system implementation.

In contrast, the baseline method, CubeCrop, shows much lower performance. Conventional methods relying on hand-crafted parameters (e.g., CFAR factor, clustering threshold) are difficult to fine-tune and cannot adapt to varying drone reflections at different positions, resulting in many missed detections and higher false negatives. Additionally, the data-driven classifier has not effectively utilized doppler information, relying only on reflections from the RAE cube. This potentially leads to incorrect classifications than our method.

\begin{table}[htbp]
\scriptsize 
\setlength{\tabcolsep}{6.5pt} 
\centering
\begin{tabular}{@{}l|c|c|c|c|c|c@{}}
\toprule
Methods              & Overall & 0-3m & 3-6m & 6-9m & 9-12m & 12-15m \\ \midrule
mDrone - small        & 1.38    & 0.49 & 0.63 & 1.10 & 1.58  & 3.06   \\
mDrone - large       & 1.26    & 0.34 & 0.71 & 0.86 & 1.56  & 2.40   \\
PointCloud - small    & 2.17    & 0.82 & 1.23 & 2.29 & 3.34  & 4.71   \\
PointCloud - large   & 1.47    & 0.58 & 1.09 & 1.23 & 1.75  & 2.81   \\ \midrule
\textbf{Ours} - small          & \textbf{0.56}    & 0.18 & 0.38 & 0.51 & 0.75  & 1.07   \\
\textbf{Ours} - large         & \textbf{0.52}    & 0.13 & 0.40 & 0.45 & 0.62  & 0.95   \\
\textbf{Ours} - small (+large) & \textbf{0.53}    & 0.21 & 0.35 & 0.48 & 0.69  & 1.06   \\
\textbf{Ours} - large (+small) & \textbf{0.48}    & 0.13 & 0.33 & 0.39 & 0.59  & 0.99   \\ \bottomrule
\end{tabular}
\captionsetup{justification=centering, singlelinecheck=false}
\caption{Mean Localization Error (\si{\metre}) at Different Distances to Radar.}

\vspace{-0.6cm}
\end{table}

\subsection{Localization Error Results}

{\noindent\textbf{Baseline Methods:}} 
We compare our method with two baselines, mDrone \cite{mdrone} and PointCloud, in terms of localization error. As mentioned in Section II, mDrone is a deep learning approach for drone tracking, but it cannot handle detection, classification, or multiple drones. For fair comparison, we removed empty frames and trained mDrone separately for each drone type. PointCloud is generated using radar data from the RAD and RED cubes by applying conventional methods such as CFAR for object detection, and DBSCAN \cite{dbscan} to cluster these detections into objects, which can then be localized. Although PointCloud achieves the best tracking accuracy among other conventional methods in terms of runtime \cite{mdrone}, it struggles with classification. Therefore, we compared only the localization error for each drone type, excluding detection and classification. For our method, we compared localization results of drones detected by CubeDN.


{\noindent\textbf{Results and Findings:}} 
Table IV shows the mean absolute localization errors (in meters) across different methods and various distances to the radar. Our method, CubeDN outperforms both mDrone and PointCloud, with an average localization error of approximately \SI{50}{\centi\metre}, which is about one-third of mDrone’s and one-fourth of PointCloud’s errors. We evaluate CubeDN in three scenarios: mini-drone only, large-drone only, and dual-drone, confirming consistent localization accuracy across all cases.

We also find that localization error increases as the drone moves further from the radar. Specifically, the error is only around \SI{15}{\centi\metre} when the drone is within 3 meters of the radar, but it increases to about 1 meter at distances of 12-15 meters. This rise in error is due to the non-linear transformation from radar polar coordinates to Cartesian coordinates, where small errors in radar coordinate result in larger errors in 3D space, especially at longer distances. Angular errors, in particular, significantly contribute to this increased localization error.


The potential reason why mDrone performs less well, especially at longer distances, is due it mainly relies on RA and RE heatmaps, not fully utilizing all doppler information. This causes tracking to mostly rely on reflections, which are hard to track at further distances and are likely affected by background noise. Similarly, PointCloud relies on hand-crafted parameters to find the largest cluster is also significantly affected by noise, leading to less accurate results.


Overall, our method consistently achieves the lowest localization error across various ranges, delivering exceptional decimeter-level tracking accuracy at close distances. CubeDN significantly outperforms the baselines, demonstrating superior robustness and stability, while also being capable of detection and classification.

\subsection{Generalization Test}
We test CubeDN beyond the dataset, in an unseen outdoor environment, with both small and large drones flying together. Experiments show an average localization error of approximately \SI{40}{\centi\metre}, with only \SI{12}{\centi\metre} at distances within 3 meters, while maintaining $94\%$ AP and $85\%$ AR, demonstrating robustness to environmental changes.

Although the model was trained on a dataset with only outdoor scenarios, we also apply it to indoor scenarios in a 7-meter-long office. The results show an average localization error of approximately \SI{40}{\centi\metre}, with $86\%$ AP and $70\%$ AR. The lower AR in the indoor environment is likely due to complicated indoor conditions, such as reflections from static objects and multi-path noise, which make detection more challenging. With the help of methods like transfer learning, the trained model can potentially be adapted to more scenarios and achieve even better results.


\section{Conclusion and Future Work}
In this paper, we propose a learning-based real-time mmWave radar object detection framework, CubeDN, for tracking and classifying small and large drones. Results show that our system outperforms other data-driven and conventional detection methods in detection and localization. CubeDN demonstrates potential for expansion into more general and complex detection tasks in 3D space. Future work includes exploring the use of multiple frames as time-series data to improve detection performance and addressing more challenging scenarios, including cluttered and dynamic environments, or when mmWave sensors are in motion.





\newpage


\bibliographystyle{IEEEtran}
\bibliography{ref}

\end{document}